\documentclass[11pt,letterpaper]{article}
\usepackage{emnlp2016}
\usepackage{times}
\usepackage{latexsym}

\usepackage{amsmath}
\usepackage{caption}
\usepackage{subcaption}
\usepackage{graphicx}
\usepackage{booktabs}
\usepackage{pbox}

\emnlpfinalcopy

\def\emnlppaperid{491}

\title{Neural Machine Translation with Recurrent Attention Modeling}

\author{Zichao Yang, Zhiting Hu, Yuntian Deng, Chris Dyer, Alex
  Smola \\
         Carnegie Mellon University \\
         {\tt \{zichaoy,zhitingh,yuntiand,cdyer\}@cs.cmu.edu} \\
{\tt alex@smola.org}}

\date{}

\begin{document}

\maketitle

\begin{abstract}
  Knowing which words have been attended to in previous time steps while
  generating a translation is a rich source of information for predicting what
  words will be attended to in the future. We improve upon the attention model
  of Bahdanau et al. (2014) by explicitly modeling the relationship between
  previous and subsequent attention levels for each word using one recurrent
  network per input word. This architecture easily captures informative
  features, such as fertility and regularities in relative distortion. In
  experiments, we show our parameterization of attention improves translation
  quality.
\end{abstract}

\section{Introduction}
In contrast to earlier approaches to neural machine translation (NMT) that used
a fixed vector representation of the input
\cite{sutskever2014sequence,kalchbrenner:2013}, attention mechanisms provide an
evolving view of the input sentence as the output is generated
\cite{bahdanau2014neural}. Although attention is an intuitively appealing
concept and has been proven in practice, existing models of attention use
content-based addressing and have made only limited use of the historical
attention masks. However, lessons from better word alignment priors in latent
variable translation models suggests value for modeling attention independent
of content.

A challenge in modeling dependencies between previous and subsequent attention
decisions is that source sentences are of different lengths, so we need models
that can deal with variable numbers of predictions across variable
lengths. While other work has sought to address this problem
\cite{cohn2016incorporating,tu2016modeling,FengLLZ16}, these models either rely
on explicitly engineered features~\cite{cohn2016incorporating}, resort to
indirect modeling of the previous attention decisions as by looking at the
content-based RNN states that generated them~\cite{tu2016modeling}, or only
models coverage rather than coverage together with ordering
patterns~\cite{FengLLZ16}. In contrast, we propose to model the sequences of
attention levels for each word with an RNN, looking at a fixed window of
previous alignment decisions. This enables us both to learn long range
information about coverage constraints, and to deal with the fact that input
sentences can be of varying sizes.




In this paper, we propose to explicitly model the dependence between attentions
among target words. When generating a target word, we use a RNN to summarize
the attention history of each source word. The resultant summary vector is
concatenated with the context vectors to provide a representation which is able
to capture the attention history. The attention of the current target word is
determined based on the concatenated representation. Alternatively, in the
viewpoint of the memory networks framework \cite{sukhbaatar2015end}, our model
can be seen as augmenting the static encoding memory with dynamic memory which
depends on preceding source word attentions. Our method improves over plain
attentive neural models, which is demonstrated on two MT data sets.


\section{Model}
\subsection{Neural Machine Translation}
NMT directly models the condition probability $p(y|x)$ of target sequence
$y=\{y_1, ..., y_T\}$ given source sequence $x = \{x_1, ..., x_S\}$, where
$x_i, y_j$ are tokens in source sequence and target sequence respectively.
\newcite{sutskever2014sequence} and \newcite{bahdanau2014neural} are
slightly different in choosing the encoder and decoder network. Here we choose
the RNNSearch model from \cite{bahdanau2014neural} as our baseline model. We
make several modifications to the RNNSearch model as we find empirically that
these modification lead to better results.

\subsubsection{Encoder}
We use bidirectional LSTMs to encode the source sentences. Given a source
sentence $\{x_1, ..., x_S\}$, we embed the words into vectors through an
embedding matrix $W_S$, the vector of $i$-th word is $W_Sx_i$. We get the
annotations of word $i$ by summarizing the information of neighboring words
using bidirectional LSTMs:
\begin{align}
    \overrightarrow{h}_i =& \overrightarrow{\text{LSTM}}(\overrightarrow{h_{i-1}}, W_Sx_i) \\
    \overleftarrow{h}_i =& \overleftarrow{\text{LSTM}}(\overleftarrow{h_{i+1}}, W_Sx_i)
\end{align}
The forward and backward annotation are concatenated to get the annotation of
word $i$ as $h_i = [\overrightarrow{h}_i, \overleftarrow{h}_i]$. All the
annotations of the source words form a context set, $C = \{h_1, ..., h_S\}$,
conditioned on which we generate the target sentence. $C$ can also be seen as
memory vectors which encode all the information from the source sequences.

\subsubsection{Attention based decoder}
The decoder generates one target word per time step, hence, we can decompose
the conditional probability as
\begin{align}
    \log p(y|x) = \sum_j p(y_j|y_{<j}, x).
\end{align}
For each step in the decoding process, the LSTM updates the hidden states as
\begin{align}
    s_j = \text{LSTM}(s_{j-1}, W_Ty_{j-1}, c_{j-1}).
    \label{eq:decoder}
\end{align}
The attention mechanism is used to select the most relevant source context vector,
\begin{align}
    e_{ij} =& v_a^T\tanh(W_ah_i + U_as_j), \\
    \alpha_{ij} =& \frac{\exp(e_{ij})}{\sum_i \exp(e_{ij})},\\
    c_j =& \sum_i \alpha_{ij} h_i.
\end{align}
This can also seen as memory addressing and reading process. Content based
addressing is used to get weights $\alpha_{ij}$. The decoder then reads the
memory as weighted average of the vectors. $c_j$ is combined with $s_j$ to
predict the $j$-th target word. In our implementation we concatenate them and
then use one layer MLP to predict the target word:
\begin{align}
    \tilde{s}_j =& \tanh(W_1[s_j, c_j] + b_1) \\
    p_j =& \text{softmax}(W_2\tilde{s}_j)
\end{align}

We feed $c_{j}$ to the next step, this explains the $c_{j-1}$ term in
Eq.~\ref{eq:decoder}.

The above attention mechanism follows that of \newcite{vinyals2015grammar}.
Similar approach has been used in \cite{luong2015effective}. This is slightly
different from the attention mechanism used in \cite{bahdanau2014neural}, we
find empirically it works better.

One major limitation is that attention at each time step is not directly
dependent of each other. However, in machine translation, the next word to
attend to highly depends on previous steps, neighboring words are more likely
to be selected in next time step. This above attention mechanism fails to
capture these important characteristics and encoding this in the LSTM can
be expensive. In the following, we attach a dynamic memory vector to the
original static memory $h_i$, to keep track of how many times this word has
been attended to and whether the neighboring words are selected at previous
time steps, the information, together with $h_i$, is used to predict the next
word to select.

\begin{figure}[!htp]
  \centering
  \includegraphics[width=0.35\textwidth]{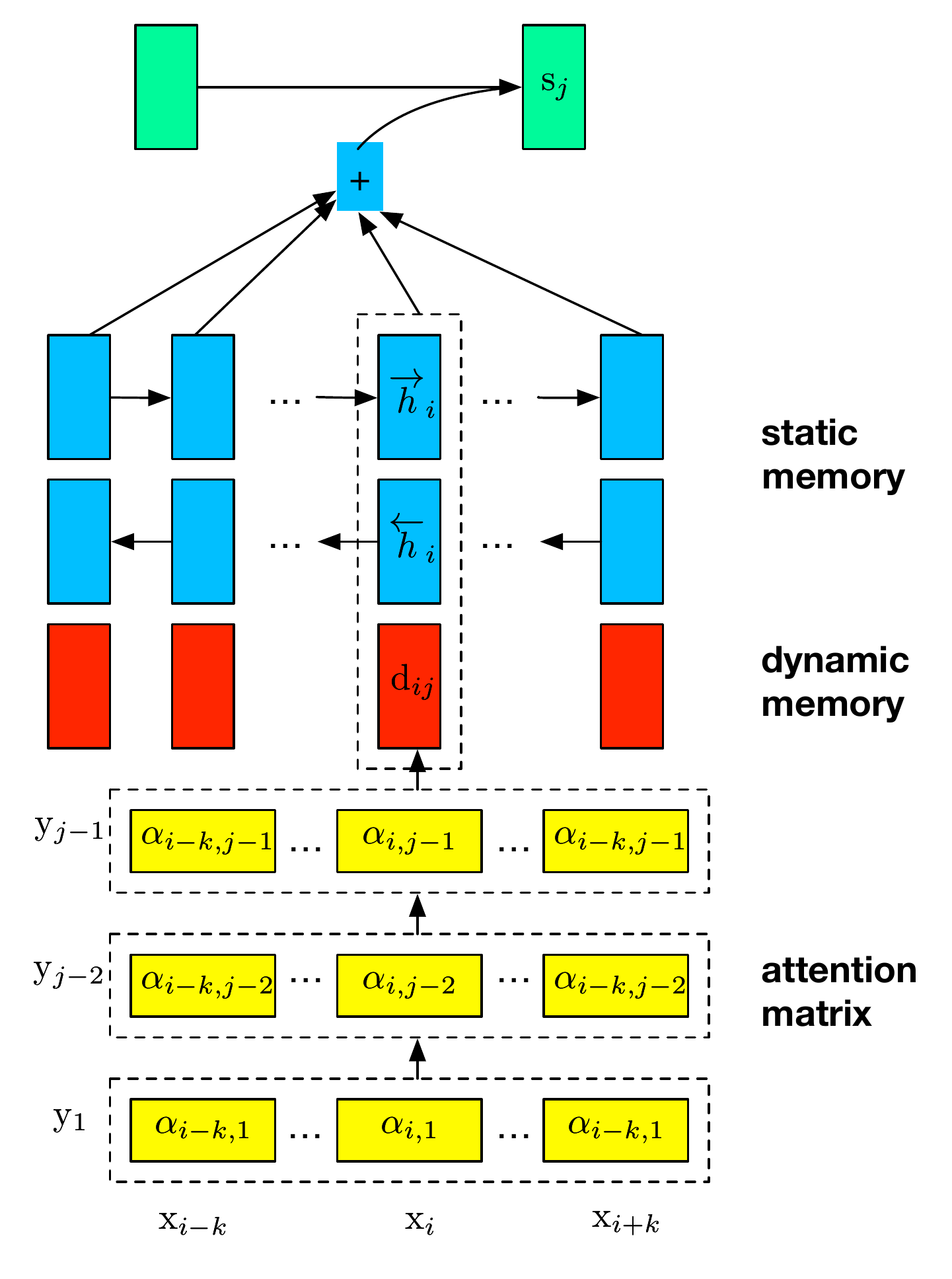}
  \caption{Model diagram} \label{fig:model}
\end{figure}

\subsection{Dynamic Memory}
For each source word $i$, we attach a dynamic memory vector $d_i$ to keep track
of history attention maps. Let
$\tilde{\alpha}_{ij} = [\alpha_{i-k, j},...\alpha_{i+k, j}]$ be a vector of
length $2k+1$ that centers at position $i$, this vector keeps track of the
attention maps status around word $i$, the dynamic memory $d_{ij}$ is updated
as follows:
\begin{align}
d_{ij}  = \text{LSTM}(d_{i, j-1}, \tilde{\alpha}_{i,j}), \forall i
\end{align}

The model is shown in Fig.~\ref{fig:model}. We call the vector $d_{ij}$ dynamic
memory because at each decoding step, the memory is updated while $h_i$ is
static. $d_{ij}$ is assumed to keep track of the history attention status
around word $i$. We concatenate the $d_{ij}$ with $h_i$ in the addressing and
the attention weight vector is calculated as:
\begin{align}
    e_{ij} =& v_a^T\tanh(W_a[h_i, d_{ij}] + U_as_j), \\
    \alpha_{ij} =& \frac{\exp(e_{ij})}{\sum_i \exp(e_{ij})},\\
    c_j =& \sum_i \alpha_{ij} h_i.
\end{align}
Note that we only used dynamic memory $d_{ij}$ in the addressing process, the
actual memory $c_j$ that we read does not include $d_{ij}$. We also tried to
get the $d_{ij}$ through a fully connected layer or a convolutional layer. We
find empirically LSTM works best.

\section{Experiments \& Results}
\subsection{Data sets}
We experiment with two data sets: WMT English-German and NIST Chinese-English.
\begin{itemize}
\item{\bf English-German} The German-English data set contains Europarl, Common
  Crawl and News Commentary corpus. We remove the sentence pairs that are not
  German or English in a similar way as in \cite{JeanCMB14}. There are about
  4.5 million sentence pairs after preprocessing. We use newstest2013 set as
  validation and newstest2014, newstest2015 as test.
\item{\bf Chinese-English} We use FIBS and LDC2004T08 Hong Kong News data set
  for Chinese-English translation. There are about 1.5 million sentences
  pairs. We use MT 02, 03 as validation and MT 05 as test.
\end{itemize}
For both data sets, we tokenize the text with
\texttt{tokenizer.perl}. Translation quality is evaluated in terms of
tokenized BLEU scores with \texttt{multi-bleu.perl}.

\begin{table*}[!thbp]
\centering
\scriptsize
  \begin{tabular}{r  p{15cm}}
    \toprule
    src & She was spotted three days later by a dog walker trapped in the quarry  \\
    ref & Drei Tage sp\"ater wurde sie von einem Spazierg\"anger im Steinbruch in ihrer misslichen Lage entdeckt\\
    baseline & Sie wurde drei Tage sp\"ater von einem Hund entdeckt . \\
    our model & Drei Tage sp\"ater wurde sie von einem Hund im Steinbruch
                gefangen entdeckt . \\
    \midrule
src & At the Metropolitan Transportation Commission in the San Francisco Bay
      Area , officials say Congress could very simply deal with the bankrupt
      Highway Trust Fund by raising gas taxes . \\
ref & Bei der Metropolitan Transportation Commission f\"ur das Gebiet der San
      Francisco Bay erkl\"arten Beamte , der Kongress k\"onne das Problem des
      bankrotten Highway Trust Fund einfach durch Erh\"ohung der Kraftstoffsteuer
      lösen . \\
baseline & Bei der Metropolitan im San Francisco Bay Area sagen
           offizielle Vertreter des Kongresses ganz einfach den Konkurs Highway
           durch Steuererh\"ohungen . \\
our model &  Bei der Metropolitan Transportation Commission in San
            Francisco Bay Area sagen Beamte , dass der Kongress durch
            Steuererh\"ohungen ganz einfach mit dem Konkurs Highway Trust Fund
            umgehen k\"onnte . \\

    \bottomrule
  \end{tabular}
  \caption{English-German translation examples}
  \label{tab:example}
\end{table*}

\subsection{Experiments configuration}
We exclude the sentences that are longer than 50 words in training. We set the
vocabulary size to be 50k and 30k for English-German and Chinese-English. The
words that do not appear in the vocabulary are replaced with UNK.

For both RNNSearch model and our model, we set the word embedding size and LSTM
dimension size to be 1000, the dynamic memory vector $d_{ij}$ size is
500. Following \cite{sutskever2014sequence}, we initialize all parameters
uniformly within range [-0.1, 0.1]. We use plain SGD to train the model and set
the batch size to be 128. We rescale the gradient whenever its norm is greater
than 3. We use an initial learning rate of 0.7. For English-German, we start to
halve the learning rate every epoch after training for 8 epochs. We train the
model for a total of 12 epochs. For Chinese-English, we start to halve the
learning rate every two epochs after training for 10 epochs. We train the model
for a total of 18 epochs.

To investigate the effect of window size $2k+1$, we report results for
$k = 0, 5$, i.e., windows of size $1, 11$.

\subsection{Results}

\begin{table}[!thbp]
\centering
  \begin{tabular}{p{5cm} c c}
    Model & test1 & test2 \\
    \toprule
    RNNSearch & 19.0 & 21.3 \\
    RNNSearch + UNK replace & 21.6 & 24.3 \\
    \midrule
    RNNSearch + window 1 & 18.9 & 21.4 \\
    RNNSearch + window 11 & 19.5 & 22.0 \\
    RNNSearch + window 11 + UNK replace & {\bf 22.1} & {\bf 25.0} \\
    \midrule
    \midrule
    \bf{\cite{JeanCMB14}} \\
    RNNSearch & 16.5 & - \\
    RNNSearch + UNK replace & 19.0 & - \\
    \midrule
    \bf{\cite{luong2015effective}} \\
    Four-layer LSTM + attention  & 19.0 & - \\
    Four-layer LSTM + attention + UNK replace & 20.9 & - \\
    \midrule
    {\bf RNNSearch + character} \\
    \cite{chung2016character} & 21.3 & 23.4 \\
    \cite{costa2016character} & - & 20.2 \\
    \bottomrule
  \end{tabular}
  \caption{English-German results.}
  \label{tab:en_de}
  \vspace{-0.5cm}
\end{table}
\begin{table}[!thbp]
  \centering
  \begin{tabular}{l  c c}
    Model & MT 05 \\
    \toprule
    RNNSearch & 27.3  \\
    RNNSearch + window 1 & 27.9 \\
    RNNSearch + window 11 & 28.8 \\
    RNNSearch + window 11 + UNK replace & {\bf 29.3} \\
    \bottomrule
  \end{tabular}
  \caption{Chinese-English results.}
  \label{tab:zh_en}
\end{table}

The results of English-German and Chinese-English are shown in
Table~\ref{tab:en_de} and \ref{tab:zh_en} respectively. We compare our results
with our own baseline and with results from related works if the experimental
setting are the same. From Table~\ref{tab:en_de}, we can see that adding
dependency improves RNNSearch model by 0.5 and 0.7 on newstest2014 and
newstest2015, which we denote as test1 and test2 respectively. Using window
size of 1, in which coverage property is considered, does not improve
much. Following \cite{JeanCMB14,luong2014addressing}, we replace the UNK token
with the most probable target words and get BLEU score of 22.1 and 25.0 on the
two data sets respectively. We compare our results with related works,
including \cite{luong2015effective}, which uses four layer LSTM and local
attention mechanism, and \cite{costa2016character,chung2016character}, which
uses character based encoding, we can see that our model outperform the best of
them by 0.8 and 1.6 BLEU score respectively. Table~\ref{tab:example} shows some
English-German translation examples. We can see the model with dependent
attention can pick the right part to translate better and has better
translation quality.

The improvement is more obvious for Chinese-English. Even only considering
coverage property improves by 0.6. Using a window size of 11 improves by
1.5. Further using UNK replacement trick improves the BLEU score by 0.5, this
improvement is not as significant as English-German data set, this is because
English and German share lots of words which Chinese and English don't.

\section{Conclusions \& Future Work}
In this paper, we proposed a new attention mechanism that explicitly takes the
attention history into consideration when generating the attention map.  Our
work is motivated by the intuition that in attention based NMT, the next word
to attend is highly dependent on the previous steps.  We use a recurrent neural
network to summarize the preceding attentions which could impact the attention
of the current decoding attention.  The experiments on two MT data sets show
that our method outperforms previous independent attentive models.  We also
find that using a larger context attention window would result in a better
performance.

For future directions of our work, from the static-dynamic memory view, we
would like to explore extending the model to a fully dynamic memory model where
we directly update the representations for source words using the attention
history when we generate each target word.

\bibliography{emnlp2016}
\bibliographystyle{emnlp2016}

\end{document}